\documentclass{article}
\usepackage{ijcai24}

\usepackage{times}
\usepackage{soul}
\usepackage{url}
\usepackage[hidelinks]{hyperref}
\usepackage[utf8]{inputenc}
\usepackage[small]{caption}
\usepackage[ruled,vlined,linesnumbered,noresetcount]{algorithm2e}
\usepackage{graphicx}
\usepackage{amsmath,amsfonts,enumitem}
\usepackage{multirow}
\usepackage{amssymb} 
\usepackage{subfigure}
\usepackage{booktabs}
\usepackage{algorithmic}
\usepackage[switch]{lineno}
\usepackage{xcolor}

\title{Graph Continual Learning with Debiased Lossless Memory Replay}

\author{
Chaoxi Niu$^1$
\and
Guansong Pang$^2$\and
Ling Chen$^{1}$
\affiliations
$^1$University of Technology Sydney\\
$^2$Singapore Management University\\
\emails
Chaoxi.Niu@student.uts.edu.au, pangguansong@gmail.com, ling.chen@uts.edu.au}

\begin{document}

\maketitle

\begin{abstract}
Real-life graph data often expands continually, rendering the learning of graph neural networks (GNNs) on static graph data impractical. Graph continual learning (GCL) tackles this problem by continually adapting GNNs to the expanded graph of the current task while maintaining the performance over the graph of previous tasks. Memory replay-based methods, which aim to replay data of previous tasks when learning new tasks, have been explored as one principled approach to mitigate the forgetting of the knowledge learned from the previous tasks. In this paper we extend this methodology with a novel framework, called Debiased Lossless Memory replay (DeLoMe). Unlike existing methods that \textit{sample nodes/edges of previous graphs} to construct the memory, DeLoMe \textit{learns small lossless synthetic node representations} as the memory. The learned memory can not only preserve the graph data privacy but also capture the holistic graph information, for which the sampling-based methods are not viable. Further, prior methods suffer from bias toward the current task due to the data imbalance between the classes in the memory data and the current data. A debiased GCL loss function is devised in DeLoMe to effectively alleviate this bias. Extensive experiments on four graph datasets show the effectiveness of DeLoMe under both class- and task-incremental learning settings.
\end{abstract}

\section{Introduction}
Due to the superior capacity to represent complex relations between samples, graph is widely used in various real-world applications \cite{xia2021survey,wu2020comprehensive} such as social networks, citation networks, and online shopping. For example, in the context of online shopping, consumers could be nodes and the edges between consumers could represent that they have purchased or rated the same product. In real-world applications, graph data are often expanded continually, e.g., new consumers and the associated connections would be constantly added to the online shopping network. Following the message propagation paradigm, graph neural networks (GNNs) \cite{wu2019simplifying,kipf2016semi} have achieved remarkable success for various graph-related tasks. Despite the success, most GNNs operate on static graph data. Directly applying them to accommodate new emerging graphs would cause the GNNs to easily forget the knowledge learned from the previous data due to the large distribution difference between the historical data and the newly added data. The forgetting of the learned knowledge when learning new graphs, a.k.a. catastrophic forgetting, would result in deteriorated performance on historical graphs. A simple solution is to store all historical data and repeatedly retrain the GNNs whenever the graph is updated, but it is prohibitively expensive in terms of computational time and resources considering the large scale of the continually expanding graph. Moreover, the previous graph data would also be inaccessible in privacy-critical application scenarios when learning the newly added data. 

To tackle catastrophic forgetting, many methods have been proposed and demonstrated impressive performance on Euclidean data such as images and texts~\cite{hayes2020lifelong,wu2021striking,wang2023comprehensive}. However, it is ineffective to directly adopt them for graph continual learning (GCL) as graphs are non-Euclidean data that contain complex relations among a large number of nodes. 

\begin{figure}
    \centering
    \includegraphics[width=0.42\textwidth]{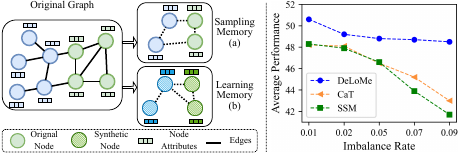}
    \caption{\textit{\textbf{Left:}} (a) Current replay-based methods use a sampling-based memory consisting of partial sampled graph data,  (b) whereas our approach learns to generate the memory using a lossless small graph with synthetic node representations. \textit{\textbf{Right:}} Average accuracy (AA) of three replay methods -- SSM, CaT and our DeLoMe -- with increasing imbalance rates on a real-world dataset ArXiv.}
    \label{motivation}
\end{figure}

Recently, to address the unique challenges in graph data, several GCL works \cite{he2023dynamically,zhou2021overcoming,liu2021overcoming,sun2023self,wang2022lifelong,evolve,rakaraddi2022reinforced,zhangsigir23,perini2022learning,zhang2022sparsified,zhang2023ricci} have been proposed to continually adapt GNNs to the expanded graph data of the current task while maintaining the performance over the graph data of previous tasks. Generally, these methods can be roughly divided into three categories, including regularization-based methods, parameter isolation-based methods, and replay-based methods. Due to the straightforward intuitiveness and impressive effectiveness, replay-based methods \cite{zhang2022sparsified,zhang2023ricci,zhou2021overcoming} have been widely explored and achieved remarkable capacity against catastrophic forgetting in GCL. Typically, they sample and store representative data of previous tasks in a memory buffer and replay them when learning a new task. However, since only selected graph information is stored, their constructed memory for each previous task often struggles to capture the \textbf{holistic graph information} of the full graph, limiting the power of memory replay against catastrophic forgetting. For example, ERGNN \cite{zhou2021overcoming} stores only individual nodes via sampling methods for replaying, which ignores the rich relations among the nodes, severely degrading the effectiveness of memory replay in GCL. A straightforward solution to this issue is to store the complete neighbors and the associated edges of the selected nodes for graph data of previous tasks, but this would pose great challenges to memory storage and is infeasible under tight memory budgets. To preserve the topological information and alleviate the storage requirement simultaneously, recent studies propose approaches like SSM \cite{zhang2023ricci} to sparsify the neighborhood of the selected nodes by filtering unimportant ones. Nevertheless, all these methods focus on constructing the memory using partial graph data, as shown in Figure~\ref{motivation}(Left), failing to preserve the holistic graph semantics. Also, these memory construction methods would become inapplicable in privacy-critical applications where the replay of previous graph data can lead to \textbf{privacy leakage}, e.g., storing the original data of the previous online shopping networks would divulge the purchase/rating information of consumers.

Further, since the amount of the memory data is often limited, the current graph data often dominates the training data, leading to a \textbf{data imbalance} between the classes in the memory data and that in the current graph data. When updating the GCL models with those imbalanced data, the models are biased toward the current task, amplifying the deteriorated performance on the previous tasks as the graph continually expands with a fixed memory budget (i.e., increasing imbalance rates), e.g., performance of SSM in Figure~\ref{motivation}(Right). 

To tackle these three issues, in this paper, we introduce a novel memory replay-based GCL approach, called Debiased Lossless Memory replay (\textbf{DeLoMe}). Rather than storing the original graph data, it learns a small graph consisting of synthetic node representations as memory so that the gradient of a randomly initialized GNN on the original large graph is lossless compared to that on the learned small graph. In doing so, the learned representations are enforced to better capture the holistic graph structure and attribute information, resulting in better performance on previous graph data, as illustrated in Figure~\ref{motivation}(Right). This node representation-based memory also helps better preserve the privacy of the graph data, compared to the original node/edge-based memory. 

To handle the aforementioned bias issue, a debiased loss function is devised in our GCL objective. This is done by calibrating the prediction logits of the classes in the memory data and the current graph data in the continual updating of our DeLoMe model, which helps largely enhance its robustness w.r.t. the class imbalance.

Our main contributions can be summarized as follows:
\begin{itemize}
    \item We propose a novel learnable memory replay-based approach DeLoMe. Compared to the concurrent work CaT \cite{liu2023cat} that introduces a seminal memory learning-based GCL method, DeLoMe introduces an enhanced graph memory learning method and augments it with a debiased GCL objective, resulting in the first GCL framework for debiased learnable memory-based replay.
    \item To obtain such memory, we introduce a lossless GCL memory learning method that utilizes gradient matching to enforce a lossless compression of large graphs of previous tasks into small synthetic graphs as the memory data. It enables DeLoMe to achieve not only a small memory budget requirement but also good privacy preservation of historical data.
    \item To mitigate the bias toward the current graph, we devise a debiased GCL objective that effectively calibrates the GNN predictions when adapting the GCL models. 
    \item Extensive experiments on four real-world datasets show that DeLoMe learns substantially more cost-effective memory and outperforms both state-of-the-art sampling and learnable memory-based methods under both class- and task-incremental settings of GCL.
\end{itemize}

\section{Related Work}
\subsection{Graph Continual Learning}
Graph continual learning (GCL) has gained growing popularity in deep learning, and various methods have been proposed \cite{he2023dynamically,zhou2021overcoming,liu2021overcoming,sun2023self,wang2022lifelong,evolve,rakaraddi2022reinforced,zhangsigir23,perini2022learning,zhang2022sparsified,zhang2023ricci}, which can be divided into three categories, i.e., regularization-based, parameter isolation-based, and data replay-based methods. For example, as a regularization-based method, TWP \cite{liu2021overcoming} preserved the important parameters in the topological aggregation and loss minimization for previous tasks via regularization terms. Among the parameter isolation methods, HPNs \cite{zhang2022hierarchical} extracted different levels of abstract knowledge in the form of prototypes and selected different combinations of parameters for different tasks, and \cite{zhangsigir23} proposed to continually expand model parameters to learn new emerging graph patterns. Differently, ERGNN \cite{zhou2021overcoming} proposed to sample and store representative nodes of previous tasks in a memory buffer, and replayed them when learning new tasks. However, the graph structure, which plays a vital role in graph representation learning, is not considered in ERGNN. To cope with the topological information, \cite{zhang2022sparsified} and \cite{zhang2023ricci} proposed the sparsification techniques to find the important neighbors of the selected nodes. Then, the important neighbors together with the edges between neighbors and representative nodes are stored in the memory buffer for replaying during the learning of new tasks. Nevertheless, all these methods focus on constructing the memory using partial graph data of previous graphs, failing to preserve the holistic graph semantics and also raising privacy concerns in privacy-critical applications. By contrast, our learned memory data helps capture the holistic graph information while effectively preserving the privacy of the previous graph data. 

\subsection{Graph Condensation}
Graph condensation aims to compress a  graph into a significantly smaller graph while preserving the holistic information of the original graph. Various graph compression methods have been proposed \cite{jin2022condensing,jin2022graph,gao2023graph,liu2022graph} that are based on gradient matching or distribution matching between the compressed graph and the original graph. In this work, we utilize gradient matching to compress the graphs with node features and structure in previous tasks into comprehensive synthetic node representations, which can serve as the memory for the subsequent replaying. Note that the concurrent work CaT \cite{liu2023cat} shares a common motivation with our method. However, unlike CaT that adopts a distribution matching-based method to learn the memory, we propose to use a gradient matching method that can measure the loss of condensation in a more fine-grained manner. CaT performs the GNNs training within the learned memory bank to alleviate the class imbalance problem, but it can lead to less effective utilization of the graph data of the current task. We instead devise a debiased GCL objective to calibrate the GCL predictions while avoiding the inefficient exploitation of the current graph data.

\section{Preliminaries}

\subsection{The GCL Problem} In this paper, we focus on the node-level GCL problem. Formally, this problem can be formulated as learning a model on a sequence of graphs (tasks) $\{ \mathcal{G}_1, \ldots, \mathcal{G}_T\}$ where $T$ is the number of continual learning tasks. Each $\mathcal{G}_t = (A_t, X_t)$ is a newly emerging graph at task $t$, where $A_t$ denotes the relations between nodes, $X_t$ represents the node features, and the labels of nodes can be denoted as $Y_t$. Generally, each task contains a unique set of classes, i.e., \{$Y_t \cap Y_j = \varnothing | t \neq j$\}. When learning task $t$, the model trained from previous tasks only has access to current data $\mathcal{G}_t$. The goal is to adapt the model to current graph $\mathcal{G}_t$ while maintaining the classification performance on the previous graphs $\{\mathcal{G}_1, \ldots, \mathcal{G}_{t-1}\}$.

\subsection{Class \& Task Incremental Settings of GCL}
Depending on whether the task indicator is provided at the testing stage, GCL can be further divided into two settings: Class-Incremental Learning (CIL) and Task-Incremental Learning (TIL). Assume that each task in $\{\mathcal{G}_1, \ldots, \mathcal{G}_T\}$ has the same number of $C$ classes and we use $C_t$ to represent the unique set of classes in $\mathcal{G}_t$. In CIL, after learning all the $T$ tasks, the model is required to classify each test instance into one of all the learned $T\times C$ classes. While under the TIL setting, the task indicator $t$ is also provided with the test instance. Thus, the model is only required to assign a class to the test instance from the classes set $C_t$ of task $t$. Compared to TIL, CIL is more practical yet more challenging as the label prediction space contains all the learned classes so far. In this paper, we evaluate the proposed method under both settings to demonstrate its effectiveness.

\subsection{Memory Replay in Graph Continual Learning}

In GCL, a GNN model is sequentially trained across the task sequence $\{\mathcal{G}_1, \ldots, \mathcal{G}_T\}$. Typically, the model only has access to $\mathcal{G}_t = (A_t, X_t)$ at task $t$. 
To continually accommodate the new graph and alleviate the catastrophic forgetting, memory replay-based methods store representative data of previous $t-1$ tasks into a memory buffer $\mathcal{B}_t$. Then, the memory data are replayed with the data of task $t$ to fit the new graph data and maintain the knowledge of previous tasks simultaneously. Therefore, the training objective of memory replay at task $t$ can be formulated as follows:
\begin{equation}\label{memoryloss}
    \mathcal{L} = \underbrace{\ell(f_{\theta}(\mathcal{G}_t), Y_t)}_{\text{current task loss}} + \lambda \underbrace{\ell(f_{\theta}(\mathcal{B}_t), Y_{\mathcal{B}_t})}_{\text{memory loss}}\, ,
\end{equation}
where $f_{\theta}(\cdot)$ is the GNN model parameterized by $\theta$, $Y_{\mathcal{B}_t}$ denote the labels of nodes in the memory buffer $\mathcal{B}_t$ (i.e., all class labels encountered in the previous $t-1$ tasks), $\ell(\cdot)$ denotes a loss function, i.e., cross-entropy loss, and $\lambda$ is a hyperparameter to control the importance of the memory loss.

The memory buffer plays an important role in maintaining previously learned knowledge and different memory construction methods have been proposed. For example, ERGNN \cite{zhou2021overcoming} sampled the representative nodes of the previous tasks as the memory. However, the rich topological information is neglected in ERGNN. Other approaches like SSM \cite{zhang2023ricci} aim to utilize the structure information by sparsifying the neighbors of the sampled nodes based on their topological importance. Then, the important neighborhood structures together with the sampled nodes are used to construct the memory. Despite the success achieved by these methods, the memory buffer constructed with partial graph data fails to preserve the holistic semantics of the original graph and it may lead to privacy leakage issue when the sampled nodes/edges are sensitive data. Further, the memory budget is typically kept significantly smaller than the current graph data, so it can lead to class imbalance between the memory data and the current graph data. Our method DeLoMe is proposed to tackle these three issues.

\begin{figure}
    \centering
    \includegraphics[width=0.49\textwidth]{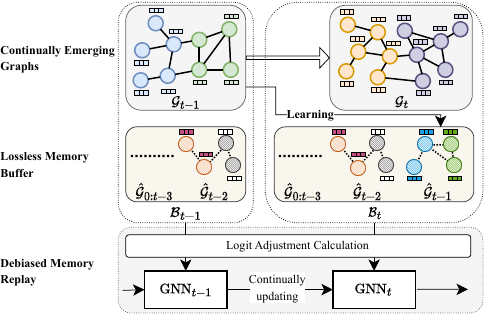}
    \caption{Overview of DeLoMe. We take two consecutive tasks ($\mathcal{G}_{t-1}$ and $\mathcal{G}_{t}$) as an example, where $\text{GNN}_{t-1}$ and $\text{GNN}_{t}$ represent the GNN model trained after the task $t-1$ and $t$ respectively. At task $t-1$, we learn synthetic node representation-based memory $\hat{\mathcal{G}}_{t-1}$ for $\mathcal{G}_{t-1}$ and add it to the memory buffer via $\mathcal{B}_t=\mathcal{B}_{t-1}\cup \hat{\mathcal{G}}_{t-1}$. At task $t$, the memory buffer $\mathcal{B}_t$ is replayed with the current graph data $\mathcal{G}_t$ to train the model $\text{GNN}_{t}$ using our debiased GCL objective. The process is repeated until all the tasks are learned.}
    \label{freamwork}
\end{figure}

\section{Debiased Lossless Memory Replay}
\subsection{Lossless Memory Learning}
\subsubsection{Key intuition}
Instead of using partial graph data as memory replay data, we propose to learn a small set of synthetic node representatives as the memory data which can holistically represent the original graph structure and attributes. Taking the task $t-1$ as an example, given $\mathcal{G}_{t-1} = (A_{t-1}, X_{t-1})$ with the label set $Y_{t-1}$, we aim to learn a \textit{compressed synthetic graph} $\hat{\mathcal{G}}_{t-1} = (I, \hat{X}_{t-1})$ associated with label set $\hat{Y}_{t-1}$ as our memory data, where the fixed identity matrix $I$ represents the structure of $\hat{\mathcal{G}}_{t-1}$. Note that the size of $\hat{X}_{t-1}$ is constrained by the memory budget and is significantly smaller than $X_{t-1}$. 

Since the learned $\hat{\mathcal{G}}_{t-1}$ captures the holistic semantics of the original graph $\mathcal{G}_{t-1}$ at task ${t-1}$, we can expect that a GNN model trained on $\hat{\mathcal{G}}_{t-1}$ would achieve comparable performance to that trained on $\mathcal{G}_{t-1}$. Following this idea, the learning objective of $\hat{\mathcal{G}}_{t-1}$ can be formulated as follows:
\begin{equation}\label{cond1}
    \min_{\hat{\mathcal{G}}_{t-1}} \ell(f_{\hat{\theta}}(\mathcal{G}_{t-1}), Y_{t-1})\, , \ \ \text{s.t.} \ \ \hat{\theta} = \arg \min_{\theta} \ell(f_{\theta}(\hat{\mathcal{G}}_{t-1}), \hat{Y}_{t-1})\, ,
\end{equation} 
where $\hat{\theta}$ denotes the parameters of the graph neural network $f(\cdot)$ trained on $\hat{\mathcal{G}}_{t-1}$. Due to the nested loop optimization of Eq.~\eqref{cond1}, directly solving the above objective would be prohibitively expensive. To address this challenge, one-step gradient matching \cite{jin2022condensing} is proposed, which aims to match the gradient of the same model with regard to the real data and the synthetic data at the first training epoch. Inspired by this, the optimization objective Eq.~\eqref{cond1} can be transformed into a lossless compression as follows:
\begin{equation}\label{cond2}
    \min_{\hat{\mathcal{G}}_{t-1}} d(\nabla_{\theta}\ell(f_{\theta}(\mathcal{G}_{t-1}), Y_{t-1}), \nabla_{\theta}\ell(f_{\theta}(\hat{\mathcal{G}}_{t-1}), \hat{Y}_{t-1}))\, ,
\end{equation}
where $d(\cdot)$ is a distance function to measure the difference between the two gradients. Moreover, to get more generalized $\hat{\mathcal{G}}_{t-1}$ without fitting to a specific model initialization, we aim to devise an objective of Eq.~\eqref{cond2} that can be minimized under different random initializations of the $f_{\theta}(\cdot)$. Thus, the final objective to learn $\hat{\mathcal{G}}_{t-1}$ is defined as follows:
\begin{equation}\label{cond3}
    \min_{\hat{\mathcal{G}}_{t-1}} \sum_{\theta_p \sim \Theta} d(\nabla_{\theta_p}\ell(f_{\theta_p}(\mathcal{G}_{t-1}), Y_{t-1}), \nabla_{\theta_p}\ell(f_{\theta_p}(\hat{\mathcal{G}}_{t-1}), \hat{Y}_{t-1}))\, ,
\end{equation}
where $\theta_p$ is a random instantiation of the parameter space $\Theta$.

By optimizing Eq.~\eqref{cond3}, we obtain the synthetic graph $\hat{\mathcal{G}}_{t-1}$ and use it to update the memory buffer $\mathcal{B}_{t-1}$, i.e., $\mathcal{B}_{t}=\mathcal{B}_{t-1}\cup\hat{\mathcal{G}}_{t-1}$. Then, $\mathcal{B}_t$ is replayed with the graph $\mathcal{G}_t$ when learning the next task $t$ (see Figure~\ref{freamwork}).

\subsubsection{The learning algorithm} 
To obtain the memory $\hat{\mathcal{G}}_{t-1} = (I, \hat{X}_{t-1})$ with $\hat{Y}_{t-1}$ of task $t-1$, we need to learn the node feature $\hat{X}_{t-1}$ and label set $\hat{Y}_{t-1}$.
Since $\hat{Y}_{t-1}$ represents the node label and is discrete, $\hat{Y}_{t-1}$ is fixed as the same classes as the original label set $Y_{t-1}$. Therefore, we only need to learn $\hat{X}_{t-1}$ for task $t-1$. To accelerate the learning process, let $b$ be the memory budget for each class, $\hat{X}_{t-1}$ is initialized as the features of randomly selected $b$ nodes from each class in $\mathcal{G}_{t-1}$. To further reduce the computation cost, we sample a fixed number of neighbors for each node in $\mathcal{G}_{t-1}$ at each hop, and adopt the mini-batch training strategy. In addition, the gradient matching and learning of $\hat{X}$ is performed on each class separately. Specifically, for a given class $c$ in $Y_{t-1}$, a batch of nodes belonging to class $c$ is randomly sampled from $\mathcal{G}_{t-1}$ together with the associated neighborhoods, which is denoted as $\mathcal{G}_{t-1}^c=(A_{t-1}^c, X_{t-1}^c)$. Meanwhile, we get the corresponding nodes of class $c$ from $\hat{\mathcal{G}}_{t-1}$ and denote it as $\hat{\mathcal{G}}_{t-1}^c = (I, \hat{X}_{t-1}^c)$. Then, the sampled $\mathcal{G}_{t-1}^c$ and $\hat{\mathcal{G}}_{t-1}^c$ are fed into the same GNN model to calculate the gradient matching loss. Finally, $\hat{X}_{t-1}^c$ is optimized via gradient descent to minimize the graph matching loss. Note that the graph neural network $f_{\theta}(\cdot)$ is not updated during the learning process and we adopt different initializations of $f_{\theta}(\cdot)$ to learn a generalized $\hat{\mathcal{G}}_{t-1}$. By imitating the training trajectory of the original graph $\mathcal{G}_{t-1}$, $\hat{\mathcal{G}}_{t-1} = (I, \hat{X}_{t-1})$ is enforced to capture the holistic semantics of $\mathcal{G}_{t-1}$ and the global topological information is implicitly incorporated into $\hat{X}_{t-1}$ (see Algorithm 1 in Appendix for the detailed algorithm).

\begin{figure}
    \centering
    \includegraphics[width=0.49\textwidth]{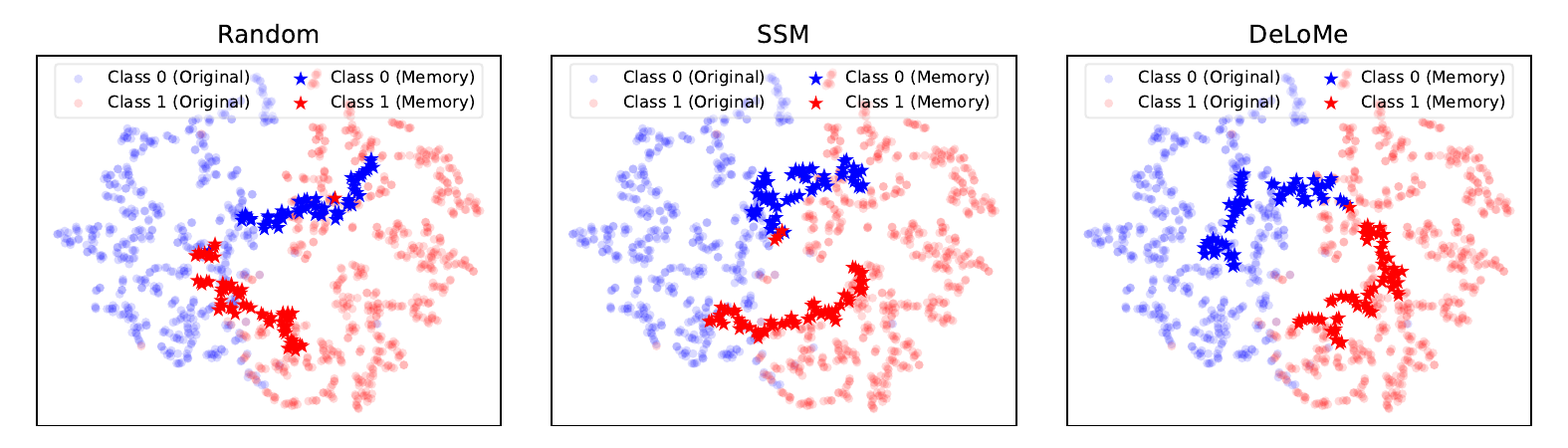}
    \caption{Visualization of node embeddings of the original graph and the memories obtained by different methods on this graph.}
    \label{vis}
\end{figure}

As illustrated in Figure \ref{vis} where we visualize the node embeddings of the memories constructed by two sampling methods (random node sampling and SSM) and DeLoMe, the synthetic node representations in the memory learned by DeLoMe can better preserve the distribution of the nodes of different classes in the original graph, compared to the other two methods. This indicates the better capability of DeLoMe in capturing more holistic graph semantics.

\subsubsection{Time complexity analysis}
We analyze the time complexity of obtaining $\hat{\mathcal{G}}_{t-1} = (I, \hat{X}_{t-1})$ at task $t-1$. As mentioned above, the learning process is conducted for each class $\mathcal{G}_{t-1}^c=(A_{t-1}^c, X_{t-1}^c)$ separately. We take a two-layer SGC \cite{wu2019simplifying} as the GNN model for gradient matching and denote the number of nodes and edges in $\mathcal{G}_{t-1}^c$ as $N^n_c$ and $N^e_c$ respectively. The dimension of the node features is denoted as $F$. For the two-layer SGC, the time complexity of forward and backward propagation with regard to $\mathcal{G}_{t-1}^c$ and $\hat{\mathcal{G}}_{t-1}^c$ are $\mathcal{O}(4N^e_cF+2N^n_cFC))$ and $\mathcal{O}(2bFC)$ respectively. Given the training epochs $E$, the time complexity for class $c$ is $\mathcal{O}(4N^e_cEF+2N^n_cEFC + 2bEFC)$. Then, the overall time complexity of learning $\hat{\mathcal{G}}_{t-1}$ is $\mathcal{O}(\sum_{c=1}^C (4N^e_cEF+2N^n_cEFC + 2bEFC))$. Since we adopt the minibatch training strategy and sample a fixed number of neighbors for each node at each hop, the numbers of $N^n_c$ and $N^e_c$ are typically small and do not induce high computation. Besides, the learning process can be implemented in parallel in practice to further reduce the learning time.

\subsection{Debiased Memory Replay}
Although the learned memory data are lossless compared to previous graphs, there is a data imbalance between the classes in the new graph $\mathcal{G}_t$ and that in the memory buffer $\mathcal{B}_t$. This is because the memory budget $b$ of each class is typically much smaller than the number of nodes belonging to new classes in $\mathcal{G}_t$. This class imbalance would increase, when the graph evolves with more current graph data or a tighter memory budget is given, amplifying the degraded performance for GCL. To tackle this problem, we propose a debiased memory replay method that adjusts the prediction logits of the classes in the memory data and the current graph data based on the class label frequencies during the memory replay. Specifically, at task $t$, we have the memory buffer $\mathcal{B}_t =\{\hat{\mathcal{G}}_1, \ldots, \hat{\mathcal{G}}_{t-1}\}$, and the vanilla objective of memory replay in Eq.~\eqref{memoryloss} can be explicitly formulated as:
\begin{equation}\label{memoryloss2}
    \mathcal{L} = \ell(H_t, Y_t) + \lambda \sum_{j = 1}^{t-1} \ell(\hat{H}_j, \hat{Y}_j)\, ,
\end{equation}
where $H_t$ and $\hat{H}_j$ are the prediction logits of $f_{\theta}(\cdot)$ for the nodes in $\mathcal{G}_t$ and $\hat{\mathcal{G}}_{j}$ respectively. Directly minimizing Eq.~\eqref{memoryloss2} would make the model biased toward the current task as the current graph data $\mathcal{G}_t$ dominates the training data. We address this problem by calibrating the logits $H_t$ and $\hat{H}_j$ based on their label frequencies. Given the memory budget $b$, the calibration magnitude for each class in the memory buffer is equal and can be defined as:
\begin{equation}\label{la1}
    \Pi_{\mathcal{B}_t} = \tau\log\frac{b}{|Y_t| + (t-1)bC}\, ,
\end{equation}
where we assume each task contains the same $C$ classes, $\tau$ is a scaling hyperparameter, and $|Y_t|$ returns the number of samples in $Y_t$. For a class $c$ in the current graph $\mathcal{G}_t$, the calibration magnitude is defined as:
\begin{equation}\label{la2}
    \Pi_t^c = \tau \log \frac{|Y_t^c|}{|Y_t| + (t-1)bC}\, ,
\end{equation}
where $|Y_t^c|$ denotes the number of training samples of class $c$ in $\mathcal{G}_t$.
By incorporating these two calibrations into Eq.~\eqref{memoryloss2}, our debiased GCL training loss is as follows:
\begin{equation}\label{finalObj}
    \mathcal{L} = \sum_{c=1}^C \ell(H_t^c + \Pi_t^c, Y_t^c) + \sum_{j = 1}^{i-1} \ell(\hat{H}_j + \Pi_{\mathcal{B}}, \hat{Y}_j)\, ,
\end{equation}
where we discard the weight parameter $\lambda$ to simplify the parameter selection. Compared to Eq.~\eqref{memoryloss2}, Eq.~\eqref{finalObj} augments the softmax cross-entropy with a pairwise margin based on label frequencies~\cite{menon2021longtail}. In this way, the predictions for dominant classes in the current graph do not overwhelm those for tail classes in the memory data, thus reducing the bias toward the dominant classes in $\mathcal{G}_t$. The detailed training steps of DeLoMe are presented in Algorithm 2 in Appendix.

\section{Experiments}

\subsection{Datasets}
Following the GCL benchmark \cite{zhang2022cglb}, four public graph datasets are employed, i.e., CoraFull \cite{mccallum2000automating}, Arxiv \cite{hu2020open}, Reddit \cite{hamilton2017representation} and Products \cite{hu2020open}. Specifically, CoraFull and Arxiv are citation networks, Reddit is constructed from Reddit posts, and Products is a co-purchasing network from Amazon. For all datasets, each task is set to contain only two classes \cite{zhang2022cglb}. Besides, for each class, the proportions of training, validation and testing are set to be 0.6, 0.2 and 0.2 respectively.

\subsection{Competing Models}
Two categories of state-of-the-art (SOTA) continual learning methods are employed for comparison. The first category contains traditional continual learning methods, i.e., EWC \cite{kirkpatrick2017overcoming}, LwF \cite{li2017learning}, GEM \cite{lopez2017gradient} and MAS \cite{aljundi2018memory}. The second category includes four SOTA GCL methods: ERGNN \cite{zhou2021overcoming}, TWP \cite{liu2021overcoming}, HPNs \cite{zhang2022hierarchical}, SSM \cite{zhang2022sparsified}, SEM \cite{zhang2023ricci} and CaT \cite{liu2023cat}. In addition, we include two other methods: Joint and Fine-tune. The Joint method is an oracle model that can see all graphs at all times and performs GCL on the full graphs of all tasks, while Fine-tune is a baseline that simply fine-tunes the learned model from previous tasks without continual learning techniques.

\subsection{Evaluation Metrics}
Average accuracy (AA) and average forgetting (AF) are adopted to evaluate the model performance. Specifically, AA and AF are calculated from the accuracy matrix $M\in \mathbb{R}^{T\times T}$, where $T$ is the number of the tasks. The entry $M_{tj} (t\geqslant j)$ denotes the classification accuracy on task $j$ after the model is optimized on task $t$. After learning all the $T$ tasks, the overall AA and AF can be calculated as follows:
\begin{equation}
    \text{AA} = \frac{\sum_{j=1}^T M_{Tj}}{T}\, , \ \ \ \text{AF} = \frac{\sum_{j=1}^{T-1} (M_{Tj} - M_{jj})}{T-1}\, .
\end{equation}

To sum up, AA evaluates the average performance of the model on all the learned tasks after learning the current task, and AF describes how the performance of previous tasks is affected by the current task. 
For both AA and AF, the higher value denotes the better GCL performance.

\subsection{Implementation Details}
To have a fair comparison, we implement the proposed method with the GCL benchmark \cite{zhang2022cglb}. More specifically, we adopt the two-layer SGC \cite{wu2019simplifying} as the backbone model with the same hyper-parameters following \cite{zhang2023ricci}. The memory budget is also set as the same in \cite{zhang2023ricci}, i.e., 60 per class for the CoraFull dataset and 400 per class for the other datasets. For the lossless memory learning module, we also use a two-layer SGC as the GNN model to match the gradients of both the original graph and the synthetic graph, and the gradient divergence is calculated based on the mean square distance. For each dataset, we report the average performance with standard deviations after 5 runs with different seeds under both task incremental and class incremental settings. 

\begin{table*}[ht]
\centering
\resizebox{0.95\textwidth}{!}{
\begin{tabular}{c|cc|cc|cc|cc}
\hline
\multirow{2}{*}{Methods} & \multicolumn{2}{c|}{CoraFull} & \multicolumn{2}{c|}{Arixv}  & \multicolumn{2}{c|}{Reddit} & \multicolumn{2}{c}{Products}\\ 
\cline{2-9} 
                    & AA/\%$\uparrow$  & AF/\%$\uparrow$  & AA/\%$\uparrow$  & AF/\%$\uparrow$ & AA/\%$\uparrow$  & AF/\%$\uparrow$ & AA/\%$\uparrow$  & AF/\%$\uparrow$ \\ 
                    
\hline \hline
Fine-tune   &3.5$\pm$0.5  &-95.2$\pm$0.5  &4.9$\pm$0.0  &-89.7$\pm$0.4  &5.9$\pm$1.2    &-97.9$\pm$3.3  &7.6$\pm$0.7  &-88.7$\pm$0.8 \\
Joint & 81.2$\pm$0.4 &-              &51.3$\pm$0.5 &-               &97.1$\pm$0.1  &-               &71.5$\pm$0.1 &-\\\hline
EWC         &52.6$\pm$8.2 &-38.5$\pm$12.1  &8.5$\pm$1.0  &-69.5$\pm$8.0  &10.3$\pm$11.6  &-33.2$\pm$26.1  &23.8$\pm$3.8 &-21.7$\pm$7.5 \\
MAS         &6.5$\pm$1.5  &-92.3$\pm$1.5  &4.8$\pm$0.4  &-72.2$\pm$4.1  &9.2$\pm$14.5   &-23.1$\pm$28.2  &16.7$\pm$4.8 &-57.0$\pm$31.9 \\
GEM         &8.4$\pm$1.1  &-88.4$\pm$1.4  &4.9$\pm$0.0  &-89.8$\pm$0.3  &11.5$\pm$5.5   &-92.4$\pm$5.9  &4.5$\pm$1.3  &-94.7$\pm$0.4 \\
LwF         &33.4$\pm$1.6 &-59.6$\pm$2.2  &9.9$\pm$12.1  &-43.6$\pm$11.9  &86.6$\pm$1.1  &-9.2$\pm$1.1   &48.2$\pm$1.6 &-18.6$\pm$1.6 \\
TWP         &62.6$\pm$2.2 &-30.6$\pm$4.3  &6.7$\pm$1.5  &-50.6$\pm$13.2  &8.0$\pm$5.2    &-18.8$\pm$9.0  &14.1$\pm$4.0 &-11.4$\pm$2.0 \\
ERGNN       &34.5$\pm$4.4 &-61.6$\pm$4.3  &21.5$\pm$5.4 &-70.0$\pm$5.5  &82.7$\pm$0.4  &-17.3$\pm$0.4  &48.3$\pm$1.2 &-45.7$\pm$1.3 \\
SSM-uniform &73.0$\pm$0.3 &-14.8$\pm$0.5  &47.1$\pm$0.5 &-11.7$\pm$1.5  &94.3$\pm$0.1  &-1.4$\pm$0.1   &62.0$\pm$1.6 &-9.9$\pm$1.3 \\
SSM-degree  &75.4$\pm$0.1 &-9.7$\pm$0.0 &48.3$\pm$0.5 &-10.7$\pm$0.3  &94.4$\pm$0.0  &-1.3$\pm$0.0   &63.3$\pm$0.1 &-9.6$\pm$0.3 \\
SEM-curvature&77.7$\pm$0.8 &-10.0$\pm$1.2  &49.9$\pm$0.6  &-8.4$\pm$1.3   &96.3$\pm$0.1   &-0.6$\pm$0.1 &65.1$\pm$1.0  &{-9.5$\pm$0.8} \\
CaT &80.4$\pm$0.5&-5.3$\pm$0.4&48.2$\pm$0.4&-12.6$\pm$0.7&97.3$\pm$0.1&-0.4$\pm$0.0&\textbf{70.3$\pm$0.9}&\textbf{-4.5$\pm$0.8}\\
% \hline
% \hline
DeLoMe (Ours)  &\textbf{81.0$\pm$0.2} &\textbf{-3.3$\pm$0.3} &\textbf{50.6$\pm$0.3}   &\textbf{5.1$\pm$0.4}  &\textbf{97.4$\pm$0.1} &\textbf{-0.1$\pm$0.1} &67.5$\pm$0.7 &-17.3$\pm$0.3\\            
\hline
\end{tabular}
}
\caption{Results (mean$\pm$std) under the class-incremental learning setting on four datasets. Fine-tune and Joint are shown to respectively serve as approximated lower bound and upper bound performance. The best performance achieved by continual learning methods on each dataset is highlighted in bold. "$\uparrow$" denotes the higher value represents better performance.}
\label{cil}
\end{table*}

\begin{table*}[ht]
\centering
\resizebox{0.95\textwidth}{!}{
\begin{tabular}{c|cc|cc|cc|cc}
\hline

\multirow{2}{*}{Methods} & \multicolumn{2}{c|}{CoraFull} & \multicolumn{2}{c|}{Arixv}  & \multicolumn{2}{c|}{Reddit} & \multicolumn{2}{c}{Products}\\ 
\cline{2-9} 
                    & AA/\%$\uparrow$  & AF/\%$\uparrow$  & AA/\%$\uparrow$  & AF/\%$\uparrow$ & AA/\%$\uparrow$  & AF/\%$\uparrow$ & AA/\%$\uparrow$  & AF/\%$\uparrow$ \\ 
                    
\hline \hline
Fine-Tune   &56.0$\pm$4.2 &-41.0$\pm$4.5 &56.2$\pm$2.6  &-36.2$\pm$2.6 & 79.5$\pm$24.2 & -11.7$\pm$4.8 & 64.4$\pm$3.8 &-31.1$\pm$4.4\\ 
Joint       &95.5$\pm$0.2 & -            &90.3$\pm$0.4  & -            & 99.5$\pm$0.0  & -             &95.3$\pm$0.8  & - \\ 
\hline
EWC         &89.8$\pm$1.0 & -5.1$\pm$0.5 & 71.5$\pm$0.6 &-0.9$\pm$0.6 & 83.9$\pm$15.1 & -2.0$\pm$1.5  &87.0$\pm$1.4  &-1.7$\pm$1.2\\
MAS         &92.2$\pm$0.9 &-3.7$\pm$1.3  &72.7$\pm$2.6  &-18.5$\pm$2.5 &61.1$\pm$7.1   &-0.5$\pm$1.0   &80.6$\pm$4.3  &-13.7$\pm$3.7\\
GEM         &91.5$\pm$0.5 &-1.9$\pm$0.9  &81.1$\pm$1.7  &-4.0$\pm$1.8      &98.9$\pm$0.1   &-0.5$\pm$0.1   &87.7$\pm$1.8  &-7.0$\pm$2.0\\
LwF         &93.8$\pm$0.1 &-0.4$\pm$0.1  &71.1$\pm$3.2  &-1.5$\pm$0.8  &98.6$\pm$0.1       &-0.0$\pm$0.0  &86.3$\pm$0.2  &-0.5$\pm$0.1\\
TWP         &94.3$\pm$0.9 &-1.6$\pm$0.4  &89.4$\pm$0.4  &0.0$\pm$0.3 &78.0$\pm$18.5  &-0.2$\pm$0.4   &81.8$\pm$3.3  &-0.3$\pm$0.8\\  
HPNs &-&-&85.8$\pm$0.7&\textbf{0.6$\pm$0.9} &-&-& 80.1$\pm$0.8 & 2.9$\pm$1.0\\
ERGNN       &86.3$\pm$1.0 &-9.2$\pm$0.9  &86.4$\pm$0.3  &0.5$\pm$0.6  &97.4$\pm$0.2   &\textbf{4.7$\pm$0.1}    &86.4$\pm$0.0  &\textbf{11.7$\pm$0.0}\\
SSM-uniform &95.3$\pm$0.5 &0.2$\pm$0.5   &88.5$\pm$0.6  &-1.3$\pm$0.5  &99.2$\pm$0.0  &-0.2$\pm$0.0   &93.1$\pm$0.8  &-1.8$\pm$0.3\\
SSM-degree  &95.8$\pm$0.3 &0.6$\pm$0.2   &88.4$\pm$0.3  &-1.1$\pm$0.1  &99.3$\pm$0.0   &-0.2$\pm$0.0   &93.2$\pm$0.7  &-1.9$\pm$0.0\\
SEM-curvature&\textbf{95.9$\pm$0.5} &0.7$\pm$0.4 &89.9$\pm$0.3 &-0.1$\pm$0.5  &99.3$\pm$0.0  &-0.2$\pm$0.0   &93.2$\pm$0.7 &-1.8$\pm$0.4\\
Cat &95.0$\pm$0.2&1.6$\pm$0.7&90.3$\pm$0.3&0.3$\pm$0.4&99.2$\pm$0.0& 0.0$\pm$0.0&94.7$\pm$0.1&-0.0$\pm$0.1\\
% \hline
DeLoMe (Ours)  &95.4$\pm$0.1 &\textbf{2.0$\pm$0.6} &\textbf{90.4$\pm$0.3} &-1.1$\pm$0.2  &\textbf{99.4$\pm$0.0}   &-0.1$\pm$0.0  &\textbf{94.8$\pm$0.1} &-2.2$\pm$0.2   \\
\hline
\end{tabular}
}
\caption{Full results (mean$\pm$std) under the task-incremental learning setting.}
\label{til}
\end{table*}

\subsection{Main Results}
\subsubsection{Results under class-incremental learning (CIL)}
The results of all methods under the CIL setting are shown in Table~\ref{cil}. Note that the results of baselines are taken from the paper \cite{zhang2023ricci} since we adopt the same GCL benchmark \cite{zhang2022cglb}\footnote{\url{https://github.com/QueuQ/CGLB/tree/master}}. From the table, we can draw the following observations: (1) Directly fine-tuning the learned model from previous tasks on the current task data leads to serious performance degradation because the knowledge of previous tasks could be easily overwritten by the new tasks. (2) Continual learning methods proposed for Euclidean data generally do not achieve satisfactory performance for GCL, which verifies the fact that the unique graph properties should be taken into consideration for GCL. (3) Replay-based graph continual learning methods achieve much better performance than other baselines. Among them, SSM and SEM outperform ERGNN on all datasets. The reason could be attributed to that SSM and SEM preserve the topological information for the historical graph data in the memory while ERGNN only stores the individual nodes. (4) Differently, CaT and DeLoMe propose to learn the memory and capture the semantics of the original graph. The performance gain of CaT and DeLoMe over SSM and SEM demonstrates that the learned memory buffer is more informative (e.g., in capturing holistic graph information) and enhances the power of replaying memory. (5) Our method DeLoMe achieves new SOTA performance in nearly all cases. Its performance can even match or outperform the ideal method Joint on the CoraFull and Reddit datasets.  Note that on Products DeLoMe outperforms all methods except CaT. This may be attributed to that the node class frequency information in Products does not help accurately capture the imbalance bias due to the possible presence of less informative nodes in this largest dataset, rendering our debiased component less effective.

\subsubsection{Results under task-incremental learning (TIL)}
In Table~\ref{til}, we report the results under the TIL setting, in which a SOTA TIL method HPNs is also included for comparison. From the table, we can first see that all the methods achieve much better performance under the TIL setting. This is because the availability of the task indicator during inference
makes TIL much easier than CIL. Despite the performance of our proposed method DeLoMe being slightly inferior to SEM \cite{zhang2023ricci} on the CoraFull dataset, DeLoMe achieves comparable performance with the oracle model, Joint. For the other three datasets, DeLoMe achieves the best performance, and even outperforms Joint on Arxiv dataset, which verifies the advantages of our two components in overcoming the forgetting and class imbalance problems.

\begin{figure}
    \centering
    \includegraphics[width=0.45\textwidth]{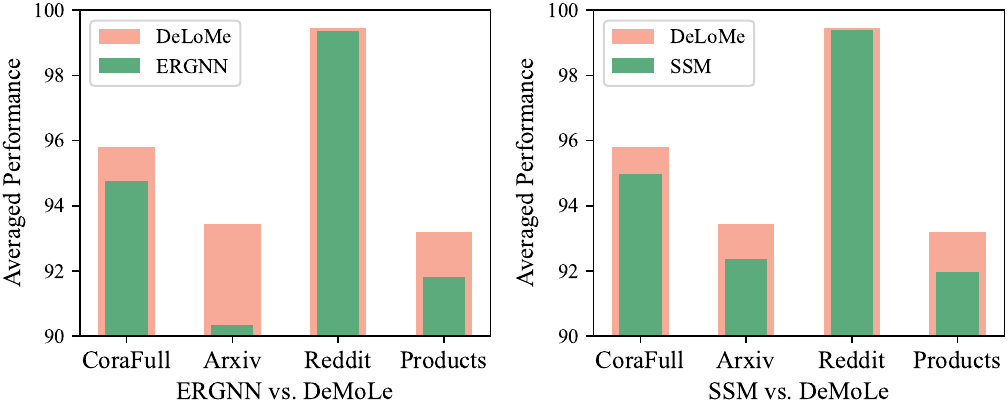}
    \caption{Average accuracy (AA) of DeLoMe against SOTA sampling-based memory construction methods on all tasks.} 
    \label{express}
\end{figure}

\subsection{Cost-effective Learning of Expressive Memory}
\subsubsection{Expressiveness of sampling- vs. learning-based memory}
In this subsection, we evaluate the expressiveness of the memory constructed by our learning-based method against two SOTA sampling-based methods, ERGNN and SSM. To evaluate the memory expressiveness, for each task, we train a GNN model with the memory data constructed by these methods and calculate the node classification accuracy on the test set of the original graph. We follow the same experimental setting in the above GCL experiments and report the average classification accuracy of all tasks in Figure~\ref{express}. It is clear that the our learning-based method achieve much better performance than both sampling-based methods, achieving improvement by up to 3\% in AA. This demonstrates the superiority of capturing the semantics of the original graph when constructing memory and explains the performance gain over the sampling-based methods in the GCL experiments.

\subsubsection{Memory budget efficiency}
\begin{figure}[htbp]
    \centering
    \subfigure[CoraFull]{
    \includegraphics[width=0.23\textwidth]{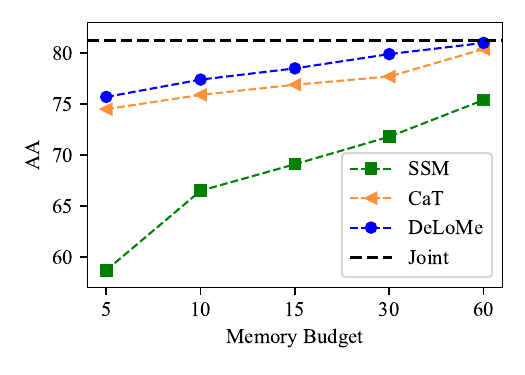}}
    \subfigure[Arxiv]{
    \includegraphics[width=0.23\textwidth]{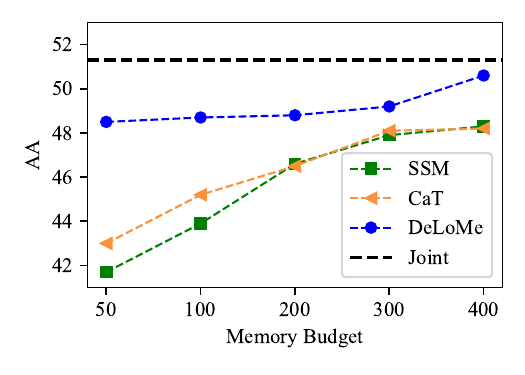}}
    \caption{AA results with different memory budgets on CoraFull and Arxiv under the class-incremental learning.}
    \label{cmemsize}
\end{figure}

The requirement of the memory budget is crucial to replay-based methods. Here we evaluate the performance of the proposed method with different memory budgets, with two best competing methods -- SSM and CaT -- and the oracle model Joint as the baselines. Due to the page limits, we report the results on CoraFull and Arxiv under class-incremental settings. The memory budgets vary in the range of $\{5, 10, 15, 30, 60\}$ for CoraFull and $\{50, 100, 200, 300, 400\}$ for Arxiv. The results are shown in Figure~\ref{cmemsize}. The figure demonstrates that the learning-based methods (DeLoMe and CaT) are more effective than the sampling-based method with tight memory budgets. Compared to SSM and CaT, our DeLoMe performs much more stably with varying budgets and achieves consistently better performance with different memory budgets, especially on the Arxiv dataset. These results reinforce our empirical evidence on the effectiveness of DeLoMe in constructing memory and handling the class imbalance problem.

\subsubsection{Computational efficiency}
We further investigate the memory construction time and inference time of DeLoMe and employ SSM and CaT for comparison. Specifically, we report the average memory construction time per task and the overall inference time of each method on the largest dataset, Products. From the results in Table~\ref{time}, we can see that CaT and DeLoMe require almost the same time for memory construction while SSM is much faster than the two learning-based methods. This is attributed to that CaT and DeLoMe involve model optimization to enhance the memory to capture the holistic semantics of the original graph while SSM employs the parameter-free sampling strategy. This also explains the superiority of CaT and DeLoMe over SSM in AA and AF in Tables \ref{cil} and \ref{til}. In terms of the inference time, the three methods are nearly the same since the memory construction only happens in the training stage and the test setting remains the same for all methods.

\begin{table}[ht]
\centering
\resizebox{0.35\textwidth}{!}{
\begin{tabular}{c|ccc}
\hline
Stage & SSM & CaT & DeloMe \\
\hline
Memory Construction &0.53&28.94&28.37\\
Inference &1.73&1.75&1.71\\
\hline
\end{tabular}}
\caption{Time (second) of different stages on Products.}
\label{time}
\end{table}

\subsection{Ablation Study}
We evaluate the contribution of two key components in DeLoMe, i.e., lossless memory learning and debiased GCL learning. Specifically, we derive four variants of DeLoMe. When the lossless memory learning is not exploited, we employ the sampling strategy to construct the memory as in ERGNN. Without loss of generality, we conduct the experiments on the CoraFull and Arxiv datasets under the CIL setting. The results of different variants are shown in the Table~\ref{ablation}. From the table, we can see that adding either of the two components contributes to a significant improvement compard to the variant that does not use both components. These showcase the importance of both components, as well as their effectiveness in respectively addressing the memory expressiveness and data imbalance problems. In general, having a more expressive memory helps achieve larger improvement than tackling the data imbalance problem. Nevertheless, with our biased GCL objective, DeLoMe can achieve further large improvement over using our expressive memory learning component alone.

\begin{table}[h]
    \centering
    \resizebox{0.35\textwidth}{!}{
    \begin{tabular}{cc|cc|cc}
    \hline
    \multirow{2}{*}{\begin{tabular}[c]{@{}l@{}} Lossless\\ Memory\end{tabular}} & \multirow{2}{*}{\begin{tabular}[c]{@{}l@{}} Debiased \\ Learning\end{tabular}} & \multicolumn{2}{c|}{CoraFull} & \multicolumn{2}{c}{Arixv}  \\
    \cline{3-6} 
    & & AA$\uparrow$ & AF$\uparrow$ & AA$\uparrow$ & AF$\uparrow$ \\
    \hline
    $\times$ & $\times$ &36.9&-59.0&24.6&-66.1\\
    $\times$ &$\checkmark$ &50.2&-40.6&33.9&-31.1\\
    % \hline
    $\checkmark$ &$\times$ &78.5 &-9.3 &47.9 &-15.8 \\
    $\checkmark$ & $\checkmark$ &81.0  &-3.3 &50.6 &5.1 \\
    \hline
    \end{tabular}}
    \caption{Ablation study of the two components in DeLoMe.}
    \label{ablation}
\end{table}

\section{Conclusion}
In this paper, we propose a novel memory replay-based GCL method DeLoMe. Traditional replay-based graph continual learning methods typically construct the memory of the previous task using partial graph data, failing to preserve the holistic semantics of the original graph at each task. To tackle this issue, we learn compressed synthetic node representations as the memory by a gradient matching approach. In this way, the learned representations capture the holistic graph structure and attribute information. Besides, the learned representations help preserve the privacy of the graph data when replaying. To overcome the class imbalance problem between the learned memory and the new-coming graph, we further proposed a debiased memory replay objective by calibrating the prediction logits of the classes in both memory data and the current task based on the label frequencies. Extensive experiments on four datasets demonstrate the effectiveness of the proposed method under both class- and task-incremental learning settings of GCL.

\bibliographystyle{named}
\bibliography{ijcai24}

\appendix

\section{Appendix}

\subsection{Algorithms}
We present the algorithm of lossless memory learning at the task $t-1$ in Algorithm \ref{alg}, and the algorithm of the proposed method DeLoMe is shown in Algorithm \ref{alg2}.

\begin{algorithm}[h]
\small
\caption{Memory Learning for graph data of task $t-1$}
\begin{algorithmic}[1]
\label{alg}
\STATE {\textbf{Input:}} Graph data $\mathcal{G}_{t-1} = (A_{t-1}, X_{t-1})$ with label $Y_{t-1}$, memory budget $b$, graph neural network $f_\theta(\cdot)$, learning rate $\eta$, and the number of epochs $E$.
\STATE {\textbf{Output:}} Memory data $\hat{\mathcal{G}}_{t-1} = (I, \hat{X}_{t-1})$ with label $\hat{Y}_{t-1}$ 
\STATE Set $\hat{Y}_{t-1}$ to fixed class values as in $Y_{t-1}$, initialize $\hat{X}$ by randomly selecting $b$ nodes from each class in $\mathcal{G}_{t-1}$.
\FOR{$e=1,\ldots, E$}
\STATE Initialize graph neural network parameter $\theta_p$ from $\Theta$\\
  \FOR{$c=1,\ldots,C$}  
  \STATE Sample $\mathcal{G}_{t-1}^c = (A_{t-1}^c, X_{t-1}^c)$ and $Y_{t-1}^c$ from $\mathcal{G}_{t-1}$
  \STATE Sample $\hat{\mathcal{G}}_{t-1}^c = (I, \hat{X}_{t-1}^c)$ and $\hat{Y}_{t-1}^c$ from $\hat{\mathcal{G}}_{t-1}$ \\
  \STATE Compute gradient: $G_{t-1}^c = \nabla_{\theta_p}\ell(f_{\theta_p}(\mathcal{G}_{t-1}^c), Y_{t-1}^c)$ 
  \STATE Compute gradient: $\hat{G}_{t-1}^c = \nabla_{\theta_p}\ell(f_{\theta_p}(\hat{\mathcal{G}}_{t-1}^c), \hat{Y}_{t-1}^c)$\\
  \STATE  Update $\hat{X} = \hat{X} -\eta \nabla_{\hat{X}} D(G_{t-1}^c, \hat{G}_{t-1}^c)$\\
  \ENDFOR
 \ENDFOR
\end{algorithmic}
\end{algorithm}

\begin{algorithm}[h]
\small
\caption{DeLoMe}
\begin{algorithmic}[1]
\label{alg2}
\STATE {\textbf{Input:}} A sequence of graph learning tasks: $\{\mathcal{G}_1, \ldots, \mathcal{G}_T\}$ and an empty memory buffer $\mathcal{B}_1$ with a budget $b$ for each class.
\STATE {\textbf{Output:}} A continually learned graph neural network $f_{\theta}(\cdot)$
\STATE Initialize $f_{\theta}(\cdot)$.
\FOR{$t=1,\ldots, T$} 
\IF{$t > 1$}
\STATE Update memory buffer $\mathcal{B}_{t-1}$ with $\hat{\mathcal{G}}_{t-1}$, i.e., $\mathcal{B}_t=\mathcal{B}_{t-1}\cup \hat{\mathcal{G}}_{t-1}$\\
\STATE Calculate the logit adjustments via Eq.(6) and Eq.(7)\\
\ENDIF
\STATE Update $f_{\theta}(\cdot)$ by minimizing training objective Eq.(8)\\
\STATE Obtain $\hat{\mathcal{G}}_t$ with Algorithm \ref{alg}\\
 \ENDFOR
\end{algorithmic}
\end{algorithm}

\subsection{Details on Datasets}
\begin{itemize}
    \item \textbf{CoraFull}\footnote{\url{https://docs.dgl.ai/en/1.1.x/generated/dgl.data.CoraFullDataset.html}}: It is a citation network containing 70 classes, where nodes represent papers and edges represent citation links between papers.

    \item \textbf{Arxiv}\footnote{\url{https://ogb.stanford.edu/docs/nodeprop/\#ogbn-arxiv}}: It is also a citation network between all Computer Science (CS) ARXIV papers indexed by MAG \cite{sinha2015overview}. Each node in Arxiv denotes a CS paper and the edge between nodes represents a citation between them. The nodes are classified into 40 subject areas. The node features are computed as the average word-embedding of all words in the title and abstract.

    \item \textbf{Reddit}\footnote{\url{https://docs.dgl.ai/en/1.1.x/generated/dgl.data.RedditDataset.html\#dgl.data.RedditDataset}}: It encompasses Reddit posts generated in September 2014, with each post classified into distinct communities or subreddits. Specifically, nodes represent individual posts, and the edges between posts exist if a user has commented on both posts. Node features are derived from various attributes, including post title, content, comments,  post score, and the number of comments.

    \item \textbf{Products}\footnote{\url{https://ogb.stanford.edu/docs/nodeprop/\#ogbn-products}}: It is an Amazon product co-purchasing network, where nodes represent products sold in Amazon and the edges between nodes indicate that the products are purchased together. The node features are constructed with the dimensionality-reduced bag-of-words of the product descriptions.
    
\end{itemize}
The statistics of these datasets are summarized in the Table \ref{datasta}.
\begin{table}[ht]
\centering
\resizebox{0.45\textwidth}{!}{
\begin{tabular}{c|cccc}
\hline
Datasets & CoraFull & Arxiv & Reddit & Products \\ 
\hline
\# nodes & 19,793 &169,343 &227,853 &2,449,028  \\ 
\# edges & 130,622 & 1,166,243 & 114,615,892 & 61,859,036 \\
\# classes & 70 & 40 & 40 & 46 \\
\# tasks & 35 & 20 & 20 & 23 \\
\# Avg. nodes per task &660 &8,467 &11,393 &122,451  \\
\# Avg. edges per task &4,354 &58,312 &5,730,794 &2,689,523\\
\hline
\end{tabular}
}
\caption{Statistics of the graph datasets.}
\label{datasta}
\end{table}

\subsection{More Implementation Details}
All the continual learning methods including the proposed method are implemented based on the graph continual learning benchmark \cite{zhang2022cglb}. For ERGNN \cite{zhou2021overcoming}, the memory budget is set to be up to 800 per class to demonstrate the advantage of the proposed method. For SSM \cite{zhang2022sparsified} and \cite{zhang2023ricci}, we conduct experiments with the same memory budgets. However, to preserve the topological information, SSM and SEM need to store the neighbors of selected nodes. The number of neighbors for each node to store is set to 5 at each hop for both SSM and SEM.

Different GNN backbones, such as GCN \cite{kipf2016semi} and SGC \cite{wu2019simplifying}, can be applied to these continual learning methods. In the main paper, to have a fair comparison to the baselines \cite{zhang2023ricci}, we employ a two-layer SGC model as the backbone. Specifically, the hidden dimension is set to 256 for all methods. The number of training epochs of each graph learning task is 200 with Adam as the optimizer and the learning rate is set to 0.005.   

For the lossless memory learning in the proposed method, we employ the same two-layer SGC model as the GNN model to calculate the gradient-matching loss. The synthetic node representations are set as learnable parameters and are initialized with the original node attributes. The labels of the synthetic nodes are fixed during the learning process. Adam is used as the optimizer to learn the synthetic representations. The learning epochs and learning rate are set to 800 and 0.0001 for all tasks and datasets respectively. The scaling parameter $\tau$ in the debiased loss function is set to 1 for all experiments for simplicity.

The code is implemented with Pytorch (version: 1.10.0), DGL (version: 0.9.1), OGB (version: 1.3.6), and Python 3.8.5. Moreover, all experiments are conducted on a Linux server with an Intel CPU (Intel Xeon E-2288G 3.7GHz) and a Nvidia GPU (Quadro RTX 6000).

\subsection{Full Results with Standard Deviation}
In the main paper, we only report the average performance of different methods. The full results with standard deviation under both class- and task-incremental settings are shown in Table \ref{cil} and Table \ref{til} respectively.

\begin{table*}[ht]
\centering
\resizebox{0.95\textwidth}{!}{
\begin{tabular}{c|cc|cc|cc|cc}
\hline

\multirow{2}{*}{Methods} & \multicolumn{2}{c|}{CoraFull} & \multicolumn{2}{c|}{Arixv}  & \multicolumn{2}{c|}{Reddit} & \multicolumn{2}{c}{Products}\\ 
\cline{2-9} 
                    & AA/\%$\uparrow$  & AF/\%$\uparrow$  & AA/\%$\uparrow$  & AF/\%$\uparrow$ & AA/\%$\uparrow$  & AF/\%$\uparrow$ & AA/\%$\uparrow$  & AF/\%$\uparrow$ \\ 
                    
\hline \hline
Fine-tune   &2.9$\pm$0.0& -94.7$\pm$0.1& 4.9$\pm$0.0& -87.0$\pm$1.5& 5.1$\pm$0.3& -94.5$\pm$2.5& 3.4$\pm$0.8& -82.5$\pm$0.8 \\
Joint       & 80.6$\pm$0.3& -& 46.4$\pm$1.4& - &99.3$\pm$0.2& - &71.5$\pm$0.7 &-\\
\hline
EWC         &15.2$\pm$0.7& -81.1$\pm$1.0& 4.9$\pm$0.0& -88.9$\pm$0.3& 10.6$\pm$1.5& -92.9$\pm$1.6& 3.3$\pm$1.2& -89.6$\pm$2.0 \\
MAS         &12.3$\pm$3.8& -83.7$\pm$4.1& 4.9$\pm$0.0& -86.8$\pm$0.6& 13.1$\pm$2.6& -35.2$\pm$3.5& 15.0$\pm$2.1& -66.3$\pm$1.5 \\
GEM         &7.9$\pm$2.7& -84.8$\pm$2.7& 4.8$\pm$0.5& -87.8$\pm$0.2& 28.4$\pm$3.5& -71.9$\pm$4.2& 5.5$\pm$0.7& -84.3$\pm$0.9 \\
LwF         &2.0$\pm$0.2& -95.0$\pm$0.2& 4.9$\pm$0.0& -87.9$\pm$1.0& 4.5$\pm$0.5& -82.1$\pm$1.0& 3.1$\pm$0.8& -85.9$\pm$1.4 \\
TWP         &20.9$\pm$3.8& -73.3$\pm$4.1& 4.9$\pm$0.0& -89.0$\pm$0.4& 13.5$\pm$2.6& -89.7$\pm$2.7& 3.0$\pm$0.7& -89.7$\pm$1.0 \\
ERGNN       &3.0$\pm$0.1& -93.8$\pm$0.5& 30.3$\pm$1.5& -54.0$\pm$1.3& 88.5$\pm$2.3& -10.8$\pm$2.4& 24.5$\pm$1.9& -67.4$\pm$1.9 \\
SSM-uniform &72.3$\pm$0.8& -15.5$\pm$1.5& 45.1$\pm$1.1& -12.2$\pm$1.4& 93.8$\pm$0.4& -2.0$\pm$0.3& 61.8$\pm$1.2& -10.7$\pm$1.1 \\
SSM-degree  &74.4$\pm$0.2& -9.9$\pm$0.1& 46.0$\pm$0.4& -11.3$\pm$0.8& 94.0$\pm$0.2& -1.9$\pm$0.2& 62.9$\pm$0.4& -10.3$\pm$0.5 \\
SEM-curvature&79.6$\pm$0.3& -2.7$\pm$0.1& 51.0$\pm$0.3& -6.7$\pm$0.6& 95.1$\pm$0.6& -1.5$\pm$0.7& 64.0$\pm$1.6& -11.2$\pm$1.8 \\
CaT &79.5$\pm$0.4&-5.5$\pm$0.2&47.1$\pm$0.4&-13.7$\pm$0.2&98.0$\pm$0.1&-0.1$\pm$0.1&\textbf{70.6$\pm$1.0}&\textbf{-4.0$\pm$0.9}\\
DeLoMe (Ours)  &\textbf{81.8$\pm$0.3} &\textbf{1.9$\pm$0.4}&\textbf{52.8$\pm$0.3}&\textbf{0.2$\pm$0.6}&\textbf{98.1$\pm$0.1}&\textbf{0.6$\pm$0.2}&67.0$\pm$0.8&-18.3$\pm$0.4\\            
\hline
\end{tabular}
}
\caption{Results under the class-incremental learning setting on four datasets with GCN as the backbone. Fine-tune and Joint are shown to respectively serve as approximated lower bound and upper bound performance. The best performance achieved by continual learning methods on each dataset is highlighted in bold. "$\uparrow$" denotes the higher value represents better performance.}
\label{cilgcn}
\end{table*}

\begin{table*}[ht]
\centering
\resizebox{0.95\textwidth}{!}{
\begin{tabular}{c|cc|cc|cc|cc}
\hline
\multirow{2}{*}{Methods} & \multicolumn{2}{c|}{CoraFull} & \multicolumn{2}{c|}{Arixv}  & \multicolumn{2}{c|}{Reddit} & \multicolumn{2}{c}{Products}\\ 
\cline{2-9} 
& AA/\%$\uparrow$  & AF/\%$\uparrow$  & AA/\%$\uparrow$  & AF/\%$\uparrow$ & AA/\%$\uparrow$  & AF/\%$\uparrow$ & AA/\%$\uparrow$  & AF/\%$\uparrow$ \\ 
\hline \hline
Fine-Tune   &58.0$\pm$1.7& -38.4$\pm$1.8& 61.7$\pm$3.8& -28.2$\pm$3.3& 73.6$\pm$3.5& -26.9$\pm$3.5& 67.6$\pm$1.6& -25.4$\pm$1.6\\ 
Joint       &95.2$\pm$0.2& - & 90.3$\pm$0.2& -&99.4$\pm$0.1& - &91.8$\pm$0.2& - \\ 
\hline
EWC          &78.9$\pm$2.4& -15.5$\pm$2.3& 78.8$\pm$2.7& -5.0$\pm$3.1& 91.5$\pm$4.2& -8.1$\pm$4.6& 90.1$\pm$0.3& -0.7$\pm$0.3\\
MAS          &93.0$\pm$0.5& -0.6$\pm$0.2& 88.4$\pm$0.2& -0.0$\pm$0.0& 98.6$\pm$0.5& -0.1$\pm$0.1& 91.2$\pm$0.6& -0.5$\pm$0.2\\
GEM          &91.6$\pm$0.6& -1.8$\pm$0.6& 87.3$\pm$0.6& 2.8$\pm$0.3& 91.6$\pm$5.6& -8.1$\pm$5.8& 87.8$\pm$0.5& -2.9$\pm$0.5\\
LwF          &56.1$\pm$2.0& -37.5$\pm$1.8& 84.2$\pm$0.5& -3.7$\pm$0.6& 80.9$\pm$4.3& -19.1$\pm$4.6& 66.5$\pm$2.2& -26.3$\pm$2.3\\
TWP          &92.2$\pm$0.5& -0.9$\pm$0.3& 86.0$\pm$0.8& -2.8$\pm$0.8& 87.4$\pm$3.8& -12.6$\pm$4.0& 90.3$\pm$0.1& -0.5$\pm$0.1\\
ERGNN        &90.6$\pm$0.1& -3.7$\pm$0.1& 86.7$\pm$0.1& \textbf{11.4$\pm$0.9}& 98.9$\pm$0.0& -0.1$\pm$0.1& 89.0$\pm$0.4& -2.5$\pm$0.3\\  
SSM-uniform  &94.5$\pm$0.8& -0.1$\pm$0.9& 88.8$\pm$1.2& -2.1$\pm$0.7& 98.8$\pm$0.3& -0.9$\pm$0.6& 90.9$\pm$2.8& -2.2$\pm$1.0\\
SSM-degree   &94.3$\pm$1.1& 0.9$\pm$0.5& 88.1$\pm$1.3& -1.0$\pm$0.4& 99.0$\pm$0.2& -0.4$\pm$0.2& 91.1$\pm$0.9& -1.8$\pm$0.8\\
SEM-curvature&95.0$\pm$0.5& -0.2$\pm$0.8& 88.8$\pm$0.1& -1.0$\pm$0.2& 99.2$\pm$0.1& -0.1$\pm$0.3& 91.6$\pm$0.5& -1.5$\pm$0.8\\
CaT &94.3$\pm$0.2&0.5$\pm$0.5&90.2$\pm$0.2&0.2$\pm$0.1&99.2$\pm$0.0&\textbf{0.1$\pm$0.1}&\textbf{94.3$\pm$0.6}&\textbf{0.1$\pm$0.2}\\
DeLoMe (Ours)  &\textbf{95.2$\pm$0.3} &\textbf{1.3$\pm$0.1}&\textbf{90.6$\pm$0.3}& 2.3$\pm$1.1&\textbf{99.3$\pm$0.1}&-0.2$\pm$0.1& 94.2$\pm$0.1&-1.4$\pm$0.1\\
\hline
\end{tabular}
}
\caption{Results under the task-incremental learning setting with GCN as the backbone.}
\label{tilgcn}
\end{table*}

\subsection{Memory budget efficiency}
\begin{figure}[htbp]
    \centering
    \subfigure[CoraFull]{
    \includegraphics[width=0.23\textwidth]{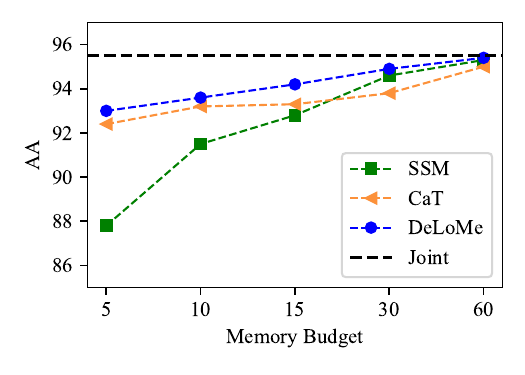}}
    \subfigure[Arxiv]{
    \includegraphics[width=0.23\textwidth]{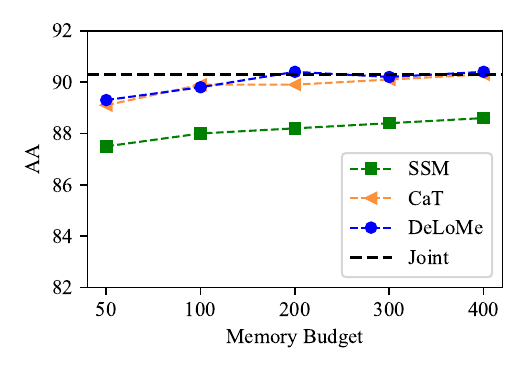}}
    \caption{AA results with different memory budgets on CoraFull and Arxiv under the task-incremental learning.}
    \label{cmemsizetl}
\end{figure}

In the main paper, we report the results of the proposed method with different memory budgets under class-incremental learning. Here, we present the results under task-incremental learning in Figure~\ref{cmemsizetl}. From the figure, we can see that all methods exhibit significantly improved performance under task-incremental learning compared to class-incremental learning. However, the learning-based methods (DeLoMe and CaT) are still more effective than the sampling-based method with tight memory budgets, indicating the importance of capturing the holistic semantics of the original graph into the memory buffer. Note that the performance gain of DeLoMe over CaT is not as significant as observed in class-incremental learning, which is attributed to the fact that task-incremental learning is less challenging than class-incremental learning.

\subsection{Results with GCN as Backbone}
As stated above, different GNNs can be used as the backbone of DeLoMe. In the main paper, we report the results with SGC \cite{wu2019simplifying} as the backbone. In this subsection, we further present the results with GCN \cite{kipf2016semi} as the backbone. Specifically, we employ the same baselines as in the main paper for comparison, and the results under the class- and task-incremental learning are shown in Table \ref{cilgcn} and Table \ref{tilgcn} respectively. From the results, we can draw similar observations as in the main paper, i.e., the proposed DeLoMe can effectively overcome the catastrophic forgetting problem in graph continual learning by utilizing lossless memory and debiased memory replay. The results also verify the effectiveness of DeLoMe with different backbones. 

\end{document}